\documentclass[times, twoside]{IEEEtran}
\usepackage{graphicx} 
\usepackage{caption}
\usepackage{amsmath}
\usepackage{makecell}
\usepackage{multirow}
\usepackage{hhline}
\usepackage[table]{xcolor}
\usepackage{pifont}
\usepackage{enumitem}
\usepackage{listings}
\usepackage{makecell}
\usepackage{float}
\usepackage{booktabs}
\usepackage{lipsum}
\usepackage{array}
\usepackage{graphicx}
\usepackage{multirow}
\usepackage{soul}
\usepackage{amsmath}    
\usepackage{svg}
\usepackage{tabularx}
\usepackage{stfloats} 
\usepackage{multicol}
\usepackage{authblk}

\usepackage{manyfoot}
\DeclareNewFootnote{R}[roman]

\date{}

\usepackage[table]{xcolor}
\usepackage[margin=1in]{geometry}
\usepackage{tabularx}
\usepackage{enumitem}

\setlist{nolistsep}

\author[ ]{Shubham Agarwal$^*$}
\author[ ]{Vlad Dinu$^*$}
\author[ ]{Thomas Searle}
\author[ ]{Mart Ratas}
\author[ ]{Anthony Shek}
\author[ ]{Dan F. Stein}
\author[ ]{James Teo}
\author[ ]{Richard Dobson}
\affil[ ]{Department of Biostatistics \& Health Informatics \protect\\ King's College London, London, U.K.}
\affil[ ]{\textit {firstname.lastname@kcl.ac.uk}}

\begin{document} 

\title{RelCAT: Advancing Extraction of Clinical Inter-Entity Relationships from Unstructured Electronic Health Records}

\maketitle

\begin{abstract}
%to be changed based on discussion section
This study introduces RelCAT (Relation Concept Annotation Toolkit), an interactive tool, library, and workflow designed to classify relations between entities extracted from clinical narratives. Building upon the CogStack MedCAT framework, RelCAT addresses the challenge of capturing complete clinical relations dispersed within text. The toolkit implements state-of-the-art machine learning models such as BERT and Llama along with proven evaluation and training methods. We demonstrate a dataset annotation tool (built within MedCATTrainer), model training, and evaluate our methodology on both openly available gold-standard and real-world UK National Health Service (NHS) hospital clinical datasets. We perform extensive experimentation and a comparative analysis of the various publicly available models with varied approaches selected for model fine-tuning. Finally, we achieve macro F1-scores of 0.977 on the gold-standard n2c2, surpassing the previous state-of-the-art performance, and achieve performance of $>=$0.93 F1 on our NHS gathered datasets.

\end {abstract}
%TC:break main
%the command above serves to have a word count for the abstract

% \begin{keywords}
% 
% \end{keywords}

\def\thefootnote{*}\footnotetext{These authors contributed equally to this work}
\section{Introduction}
\begin{figure*}[htbp]
  \centering
  \includegraphics[scale=0.5]{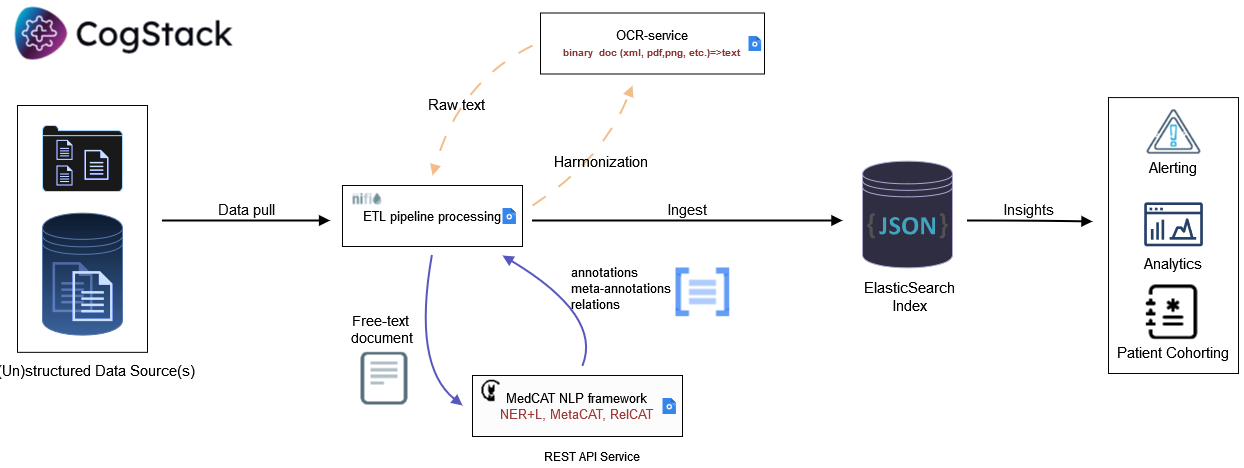}
  \caption{CogStack pipeline}
  \label{fig:cogstackpipelinefig}
\end{figure*}

In healthcare settings, unstructured clinical text holds critical insights about patient conditions, treatments, and outcomes. However, extracting meaningful data from this text is challenging due to the dispersed nature of relevant terms. Our work aims to detect relations between SNOMED CT \cite{snomed} entities that have already been extracted from unstructured clinical narratives. Our tool, named RelCAT - Relation Concept Annotation Toolkit - automatically classifies relation information between linked concepts for analytics and other workflows. An example relation would be ‘spatial’: between two concepts such as ‘left: 7771000’ and ‘lung tumour: 126713003’. The challenge here is the separation of these concepts by other words, making it difficult to capture the complete clinical concepts via traditional methods. Our work builds on previously presented work with CogStack \cite{jackson2018cogstack}, shown in figure \ref{fig:cogstackpipelinefig}, that built openly available tools for the normalisation and harmonisation of disparate clinical data sources, and MedCAT \cite{kraljevic2021multi} for the identification, linking and contextualisation of individual SNOMED CT terms from clinical narratives.  

RelCAT implements advanced natural language processing (NLP) models, and is tested on publicly available datasets alongside two real-world private datasets from two large UK National Health Service (NHS) teaching hospitals.

The toolkit offers a versatile approach to relation extraction built on advanced transformer models. It allows users to create, annotate, and fine-tune models utilising a prior published method to train, fine-tune and validate Named Entity Recognition, Linking, and Contextualisation (NER+L) models, namely MedCAT. \cite{kraljevic2021multi}. 

RelCAT improves the richness of the information extraction (IE) task by extracting relationships between medical entities such as symptoms, diagnoses, medications, and treatments. For example, extracting relationships between medications and adverse reactions enables better monitoring of potential side effects, allowing for improved care \cite{perera2020named}. Additionally, extracting spatial-entity relationships, such as the location of tumors relative to organs in radiology reports, aids in surgical planning and treatment \cite{datta2020understanding}. Supporting the extraction of relations between entities improves the richness of data extracted by information extraction toolkits such as MedCAT.

The objectives of this study are:
\begin{itemize}
    \item Relation extraction model training, evaluation and deployment framework for training, evaluating and using relation annotations within existing NER+L pipelines. 
    \item Experimental results for an open-source dataset and real-world clinical datasets from two large UK based teaching hospitals
    \item Experimental results comparing LLM-based approaches for relation extraction including in-context learning. 
\end{itemize}

\section{Methodology}
\begin{figure*}[htbp]
  \centering
  \includegraphics[scale=0.25]{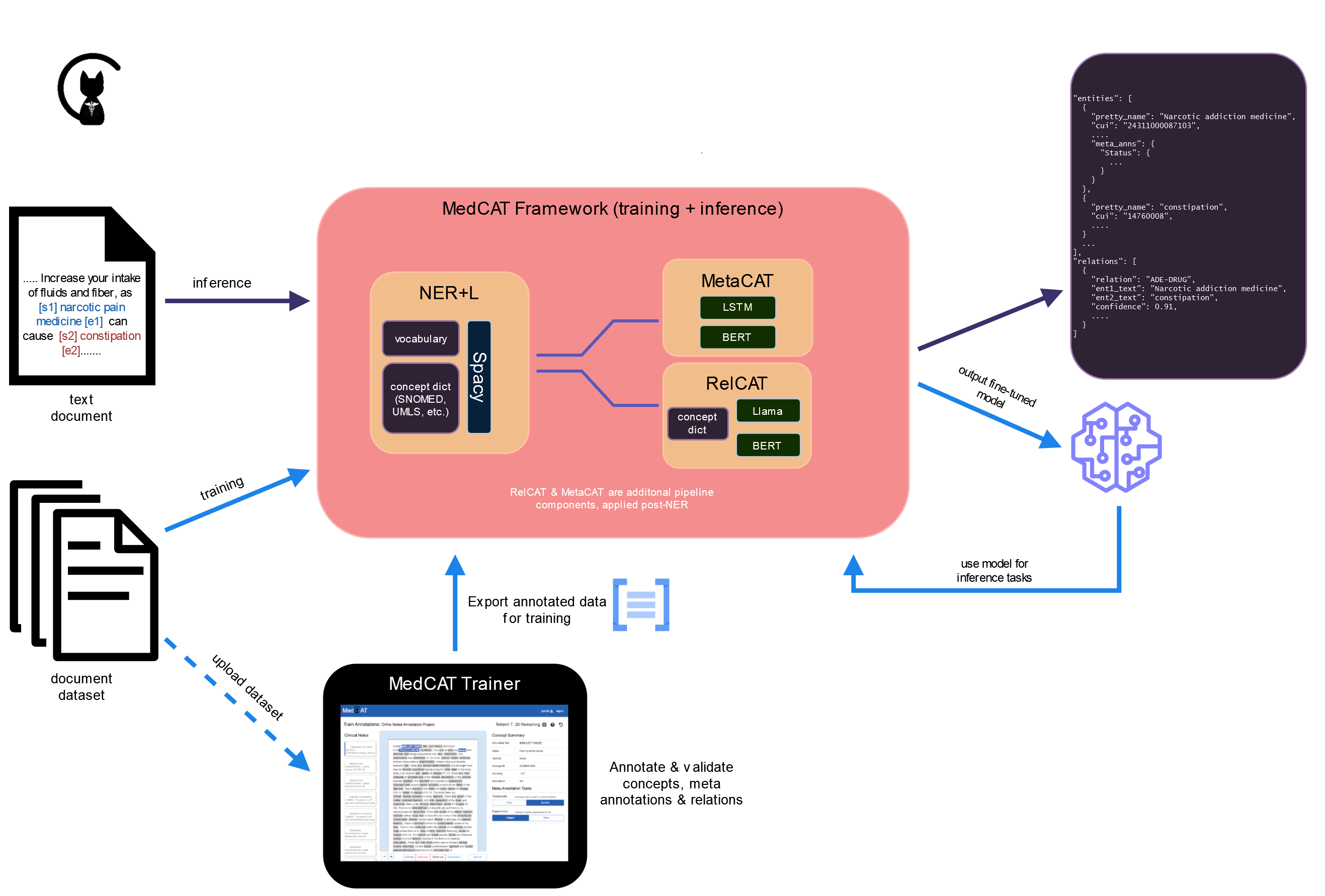}
  \caption{MedCAT NLP framework}
  \label{fig:medcatframeworkfig}
\end{figure*}

%to be changed
RelCAT, part of a complete NLP pipeline, uses MedCAT, a robust entity identification and linking solution to perform the pre-processing step\cite{kraljevic2021multi}. The pipeline tasks are: entity extraction using MedCAT (NER+L coupled with meta classification) and relation classification with RelCAT. The framework is displayed in Figure \ref{fig:medcatframeworkfig}. MedCAT's extensive, flexible and customisable vocabulary and concept database (such as the Unified Medical Language System (UMLS) \cite{journals/nar/Bodenreider04} or SNOMED-CT \cite{snomed}) enable it to extract and link diverse medical terms to their corresponding concept unique identifiers (CUIs / SCTIDs).

% The NER+L model implementation of MedCAT is based on bidirectional long short-term memory (LSTM) recurrent neural network architecture, however, in our case, we have chosen to go with BERT, as it has been shown to have equal or better performance in relation classification tasks  https://arxiv.org/pdf/2009.05451, we have also implemented the same approach with Llama for comparison as it sits on the same HuggingFace transformers library.

\subsection{Modelling Clinical Concept Relations with Deep Pre-Trained Transformer Models}\label{sec:mod-finetuning}
% BERT, based on a multi-layer bidirectional Transformer architecture \cite{devlin2018bert}, leverages pre-training to achieve remarkable performance across various natural language tasks. It often outperforms other models and, due to its compact and efficient design, is a preferred choice for tasks involving text data \cite{li2022survey} \cite{minaee2021deep}.
% LLaMA is built on an advanced auto-regressive Transformer architecture and leverages extensive pre-training on a diverse dataset to achieve high performance across a range of natural language processing tasks. Its efficiency and scalability make it a strong contender in tasks involving text classification \cite{touvron2023llama} \cite{touvron2023llama2}. 

In this work, we employ the BERT \cite{devlin2018bert} and Llama \cite{touvron2023llama} models for the relation classification task. Specifically, we utilize the pre-trained BERT model (bert-large-uncased) and Llama 3 8b model fine-tuned with two fully connected layers for classification. Both models are pre-trained with large corpora of text data, and are based upon the Transformer architecture \cite{vaswani2017attention}. 

Relation classification presents unique challenges compared to sequence or token classification, as it requires identifying relationships between two entities that may span multiple tokens. To address this, we experiment with different entity representation methods:

\begin{itemize}
\item Adding special tokens [s1], [e1], [s2], and [e2] to indicate the start and end of the entities.
\item Storing the index positions of the entities (tokenized indices) to mark their start and end points.
\end{itemize}

\begin{figure*}[htbp]
  \centering
  \includegraphics[scale=0.30]{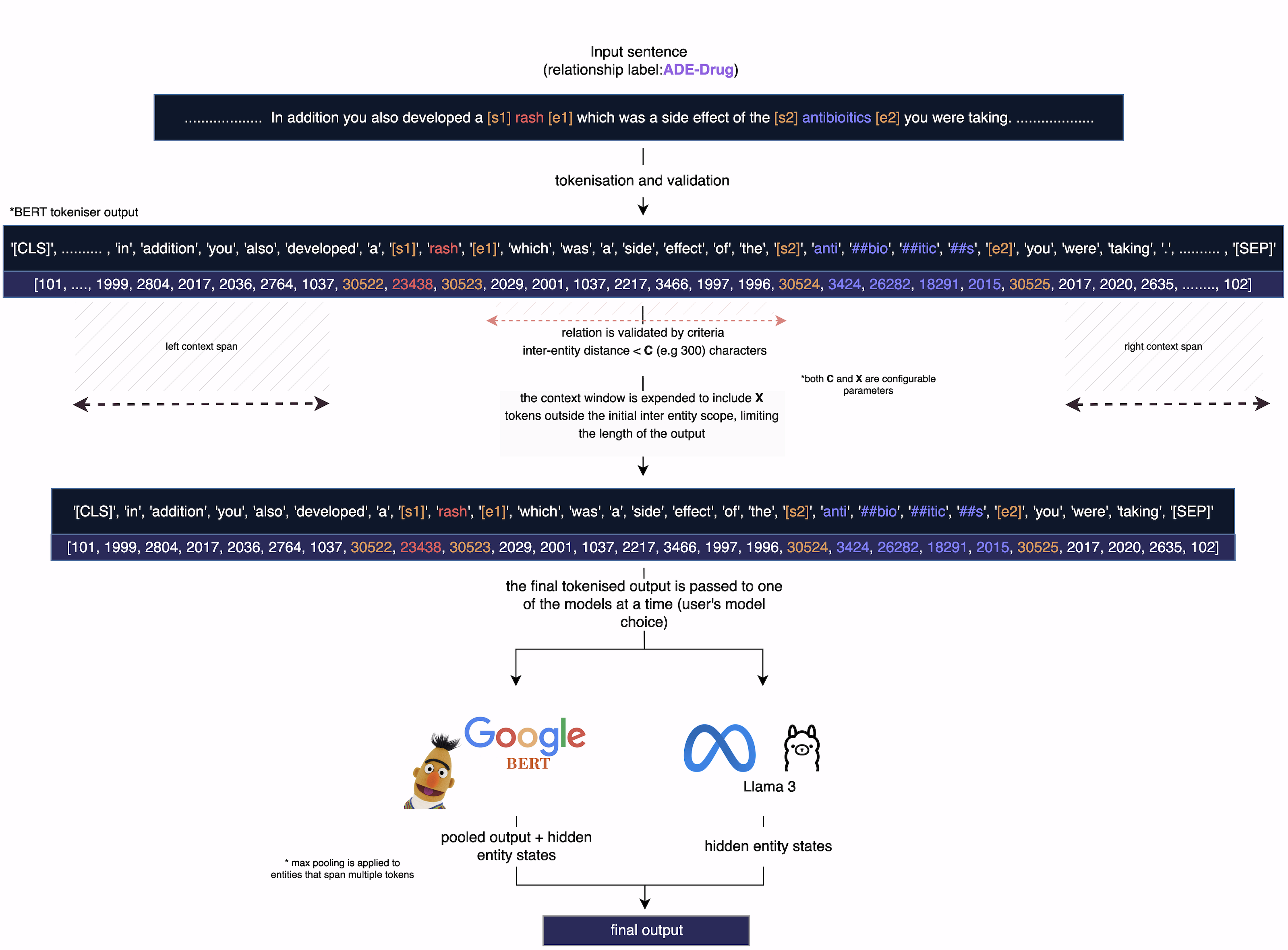}
  \caption{RelCAT processing approach}
  \label{fig:relcatprocessdiagramfig}
\end{figure*}

For both model implementations, we follow a consistent processing approach, as shown in figure \ref{fig:relcatprocessdiagramfig}:

\begin{itemize}
\item Tokenize the input text
\item Validate relation pair by considering inter entity distance (measured in characters)
\item Apply an additional context selection window, focusing on tokens close in proximity to the special tags considered 
\item Pass through an embedding model (e.g. BERT-large or Llama3.0-8B) and extract the hidden states for each entity. If an entity spans multiple tokens, apply max pooling.
\item Stack the hidden states of the entities and feed them into the fully connected layers.
\item[*] When using special tokens, we also experiment with extracting the hidden states of these tokens to provide additional information.
\end{itemize}

Moreover, after experimentation, we include the pooled output from BERT, which summarizes the entire sequence, to enhance our model's performance.

The n2c2 dataset for relation extraction reflects the class imbalance often encountered in medical records \cite{rahman2013addressing}. This imbalance comes from the rarity of some relationships compared to others. To ensure that our RelCAT performs well across all classes, we use:
\begin{itemize}
    \item \textbf{Class weights}: to modify the loss function to increase the penalization of misclassification of the minority class. Assigning higher weights to these minority class instances during training improves its ability to recognize and accurately classify them. \cite{byrd2019effect}.
    \item \textbf{Stratified batching}: Since the model does not encounter underrepresented classes as much during training, it struggles to learn effectively about these classes. Stratified batching improves performance by ensuring that each batch contains a balanced representation of all classes, unlike random batching, which can overrepresent majority classes in imbalanced datasets \cite{he2013imbalanced} \cite{afan2022linear}.
 
\end{itemize}

\subsection{Large Language Models for in-context learning}
Large Language Models (LLMs) are built on the transformer architecture \cite{vaswani2017attention}, and have set new benchmarks in natural language processing (NLP)\cite{ahuja-etal-2024-megaverse}. Examples include (now older) `encoder`-based models such as BERT, and `decoder`-based models such as Llama 3.1 8b and Mistral 7b \cite{jiang2023mistral}.
% Substantially larger than BERT's 110M parameters models the derivatives they are pre-trained on massive datasets \cite{brown2020language} and subsequent fine-tuning to follow instructions, these models exhibit remarkable proficiency in understanding and categorizing text\cite{}.

Instead of directly fine-tuning the LLMs for relation classification, as described in Section \ref{sec:mod-finetuning}, the model can be used to perform classification using zero-shot learning \cite{radford2019language} and few-shot learning \cite{wang2020generalizing}.

Prompting enables the model to leverage its pre-training to perform classification with little to no additional data needed. Zero-shot learning allows the model to directly perform inference and classify data without any specific training for the task, making it particularly valuable when labeled data are scarce. This approach is ideal for scenarios where obtaining labeled data is difficult or expensive.

Few-shot learning involves providing the model with a small set of input-output pairs, which act as reference `training data'. These examples help the model better understand the specific tasks and improve its classification accuracy. Few-shot learning strikes a balance between zero-shot learning and extensive fine-tuning by requiring only a minimal amount of labeled data to achieve significant performance gains. This approach is useful when a limited amount of labeled data is available, allowing the model to adapt and perform the classification task more effectively \cite{gao-etal-2021-making}.

\subsection{Creating and processing relations}

\subsubsection{Workflows}
RelCAT offers two distinct approaches for processing datasets in relation classification and extraction tasks:

\begin{itemize}
    \item \textbf{End-to-End Relation Extraction with MedCAT Integration}
    This workflow allows users to input raw text and receive fully extracted relations as output. Using MedCAT's NER capabilities, RelCAT can perform the entire process from entity recognition to relation extraction in one seamless operation.
    \item \textbf{Standalone Relation Classification}
    In this mode, RelCAT functions as an independent module focused solely on classifying relationships. It requires text with preidentified entities as input, making it suitable for scenarios where entity recognition has already been performed or when working with specific datasets.
\end{itemize}

\subsubsection{MedCATTrainer integration}
We adapt MedCATTrainer, an annotation tool for MedCAT models, to gather relation annotations for the two NHS datasets. The tool operates on the following principles:
\begin{itemize}
\item \textbf{Entity Annotation}

Annotators can manually identify and mark entities within the text.
Each entity is assigned a concept identifier derived from a predefined ontology dictionary, such as SNOMED CT. A user may apply CUI filters that help better define the project entity annotation scope, focusing on specific concepts if necessary. These concept identifiers are stored in a concept database (CDB).

\item \textbf{Relation Annotation}

We adapt the tool to support labeling of relationships between pairs of entities, as shown in figure \ref{fig:medcattrainerfig_rel_ann}.

\begin{figure*}[htbp]
  \centering
  \includegraphics[scale=0.25]{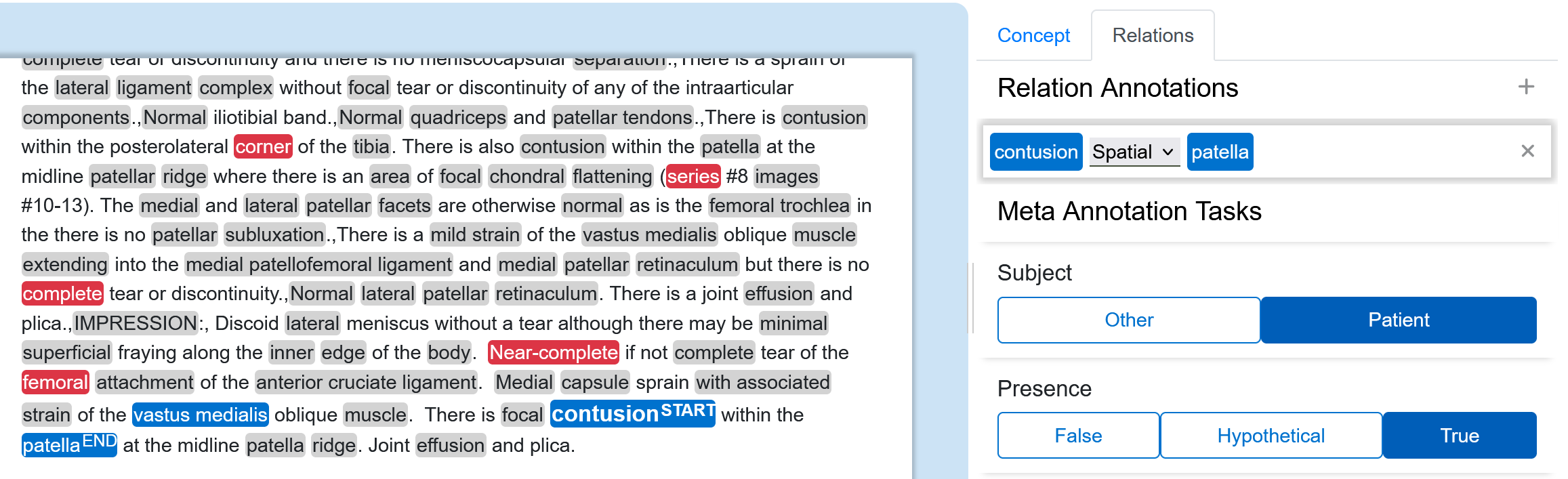}
  \caption{MedCAT Trainer interface for annotating relations}
  \label{fig:medcattrainerfig_rel_ann}
\end{figure*}

Annotators have the flexibility to determine:
\begin{itemize}
    \item the types of relations to assign
    \item the contextual span considered relevant
    \item the acceptable distance between related entities
    \item CUI filtering of the related entities
\end{itemize}

\item \textbf{User-Driven Approach}

The tool allows users to make annotation decisions.
This flexibility allows for adaptability to various text types and research needs within the medical domain.
\end{itemize}

\subsubsection{Handling non-relations}
\label{sssc:handling-non-relation}
Non-relations refer to instances where two entities in a text are not meaningfully related in the context of the task at hand. This is important because in many real-world scenarios, not all entities in a text have a significant relationship with each other. We propose enhancements such as incorporating SNOMED CUI(SCTID) knowledge and creating non-relationship types between medical entities.

It is acknowledged that defining what constitutes a non-relation can be difficult, the following challenges were identified when preparing local datasets and models:
\begin{itemize}
    \item context-dependent and may require domain expertise
%   \item balancing the dataset between relations and non-relations to avoid biasing the model
    \item ensuring that the identification of non-relations does not overshadow the detection of actual important relationships, avoiding the biasing of the model
\end{itemize}

To address some of the problems aforementioned we have implemented an automatic way of generating relations based on CUIs or type identifiers (TUIs) as a pre-training step during dataset preparation. 
It is important to note that TUIs are in essence custom-made identifiers that directly tied to the top level family of the ontological dictionary of concepts available, in the case of the NHS datasets, SNOMED-CT.
Users can specify the TUI of both relations and non-relations, meaning that as an example, a project that targets relations between "Drug" and "Adverse Effects" can be specified as a TUI pair (T01, T02), where T01 would represent the "substance" concept family and T02 "disorder", respectively.
\newline
    A prerequisite for generating non-relations and relations alike is to ensure that the considered entities have been manually validated by the annotator, this ensures that the quality of the annotations and their respective relationship type is increased, while lowering the possible occurrence of invalid entity relations.
\newline
    To ensure that the dataset is balanced, we also introduce a per project limit of possible non-relations, this option gives the user the ability to manually control the amount of samples generated. This feature has proven crucial to improving model performance after extensive testing. Furthermore, it is also possible to further split the non-relation class according to the TUI pair of each entity, this can allow for further granular control should a project be focused on relation classes that have a broader definition.

\subsection{Experimental Datasets} 
The following datasets are used for evaluation:
\vspace{1mm}
\subsubsection{n2c2 dataset}

Created as part of the National NLP Clinical Challenges 2018 \cite{henry20202018}, the dataset is comprised of annotations on 505 discharge summary notes collected from MIMIC-III (Medical Information Mart for Intensive Care-III) clinical care database \cite{10.1093/jamia/ocz166}.
We refer to this as the gold standard, as it has been validated and applied across several published studies. Table \ref{tab:n2c2dataset} describes the dataset's sample distribution across 8 classes.

\begin{table} [h]
\caption{Post-processing sample distribution for n2c2 dataset}
\label{tab:n2c2dataset}
\centering

\begin{tabular}{m{2.5cm} m{2.5cm}} 
\toprule
    Class / type & Number of relations\\
\midrule
ADE-Drug &2200\\
Route-Drug &11072\\
Form-Drug &13302\\
Frequency-Drug &12612\\
Dosage-Drug  &8438\\
Strength-Drug &13398\\
Duration-Drug &1282\\
Reason-Drug &9982\\

\end{tabular}
\end{table}

\subsubsection{NHS datasets}
Collected within UK National Health Service (NHS) hospital Foundation Trusts - Guy's and St. Thomas' \& King's College London Hospitals.

\begin{enumerate}[label=\alph*.]
\item {Spatial dataset}

\vspace{0.5mm}
The `Spatial' relation class is represented by an entity pair where the left concept is physically located in the right concept, for example: ``Multiple levels of tissue were examined, these show superficial skin shave with part of an actinic keratosis of proliferative and bowenoid type'', the relation evidenced is between ``actinic keratosis'' and ``skin''. 119 documents were annotated and varied clinical document types were used: brain MRI reports, dermatology procedures, humeral fractures, radiology CT. Entity pairs considered are directional, specifically from left to right. This means the order of the entities matters in defining the relationship. The samples considered are detailed in Table \ref{tab:spatialdataset}. A total of 613 relations were generated post-processing, the number of non-relations were limited to a maximum 70 per project to ensure a wide range of combinations is covered across, as well as to not over-sample, these relations were randomly selected. The process is described in Section \ref{sssc:handling-non-relation}. All entities encompassed are part of the SNOMED ontology dictionary, a detailed type list is provided in Table \ref{tab:spatialdatasettypes}.

\vspace{0.5mm}

\begin{table}[h]
\caption{Post-processing sample distribution for Spatial dataset}
\label{tab:spatialdataset}
\centering

\begin{tabular}{m{3cm} m{2.5cm}} 
\toprule
    Class / type & Number of relations\\
\midrule
Spatial &281\\
Other &332\\

\end{tabular}
\end{table}

\item {Physiotherapy-Mobility dataset}

This dataset was annotated at King's College Hospital. It collected mentions of physiotherapy and mobility concepts, some of which were not available in SNOMED CT, describing patients ability to carry out daily such as mobility whilst unwell as inpatients in hospital.
We define the 'Single instance' relation class as an entity pair where the left and right entities together form a single concept. An example of such a relation is : "to be able to step transfer with WZF [Wheeled Zimmer Frame] and AO1 [assistance of 1] in 1/52", the related entity pair being "WZF" and "AO1". For this example, whilst 'wheeled zimmer frame' appears in SNOMED CT, 'assistance of one' does not. This relation class was annotated in 486 documents, specifically in physiotherapy notes. The order of the entities is significant in defining this relationship, as the left entity is understood to be related to the right entity in a specific way that forms a unified concept or instance name. The number of different relation types is shown in Table \ref{tab:physiotherapydataset}. A notable difference in comparison to the 'Spatial' dataset is that some of the relations and non-relations are considered 'custom' concepts that are not part of SNOMED CT. This presents an increased level of difficulty for both training and annotating. Table \ref{tab:singleinstancedatasettypes} displays the entity type distribution by class.

\vspace{0.5mm}

\begin{table}[h]
\caption{Post-processing sample distribution for Physiotherapy-Mobility dataset}
\label{tab:physiotherapydataset}
\centering

\begin{tabular}{m{3cm} m{3cm}} 
\toprule
    Class / type & Number of relations\\
\midrule
Single Instance &278 \\ % was 183  % was 161
Other &301\\ %% was 151

\end{tabular}
\end{table}
\end{enumerate}

\vspace{0.5mm}

In both NHS datasets, the 'Other' class has been generated from the valid non-relation labeled entity pairs available whilst attempting to limit and closely match the number of samples to the true class. This procedure is described in section \ref{sssc:handling-non-relation}.

\section{Results}
This section reports the performance of the RelCAT entity relation classification models according to accuracy, macro F1-score, recall and precision. 

\vspace{1mm}
\subsubsection{Performance on n2c2 dataset} 
We conducted experiments using three configurations for the model layers: freezing all layers, unfreezing all layers, and unfreezing only the last layer. The detailed results for these configurations are presented in Tables \ref{tab:bert_n2c2}, \ref{tab:biobert_n2c2}, \ref{tab:bioclinicalbert_n2c2}, \ref{tab:bert_lastu_n2c2}, and \ref{tab:llama_n2c2}, while Table \ref{tab:perfsummaryn2c2} contains a summary of the results. Unfreezing layers enables fine-tuning of the pre-trained feature representations, allowing them to adapt to the specific characteristics of our dataset.
Due to the model's size, Llama was implemented with all layers frozen and with only the last layer unfrozen.
\begin{enumerate}[label=\alph*.]
    \item {\textbf{BERT:}} BERT model with all layers unfrozen and with the last layer unfrozen demonstrates exceptional performance across most classes, achieving recall values of 0.97 or higher, with the exception of three classes. Notably, the minority classes - ADE-Drug and Duration-Drug have high recall values of \textbf{0.86} and \textbf{0.94}, respectively. 

% This improvement can be attributed to the model's ability to grasp and understand nuances in more depth, leading to a model that is very well fitted to the data. Both variations have similar performance, although this can be due to performance being already at the ceiling.

\item{\textbf{Llama:}} With the last layer unfrozen, the Llama model demonstrated high overall performance. However, for the minority classes, the performance falls short and leaves room for further enhancement. BERT outperforms Llama both in overall performance and specifically on minority class predictions.

\item{\textbf{Variants of BERT}}:
In addition to BERT, we utilized BioBERT and BioClinicalBERT for classification tasks. 
With all layers unfrozen, both models achieve recall values above 0.9 for all classes, except the minority classes, for which the models still perform well for Duration-Drug, but struggle on the ADE-Drug class.
\end{enumerate}
BERT model with all layers unfrozen has the best performance on the n2c2 dataset, achieving a macro F1-score of 0.977.

\noindent\textbf{\textit{Comparison with state-of-the-art}:}
The original n2c2 paper \cite{henry20202018} reports the highest macro F1-score at 0.956; and a subsequent study using transformer models \cite{yang2021clinical} achieved a top F1-score of 0.9610. Our work surpasses both, with an overall F1-score of \textbf{0.977}. In terms of minority class performance, as only the original n2c2 paper reported class-specific F1-scores, our results also exceed their best-performing model, with F1-scores of \textbf{0.86} and \textbf{0.93} for ADE-Drug and Duration-Drug, compared to their 0.85 for both classes.

\bigbreak
\subsubsection{Performance on NHS datasets} 
Similar to the n2c2 dataset, multiple experiments were conducted, including the BERT large model with frozen and unfrozen layers, Llama 3 and using the fine-tuned n2c2 BERT models as a base for the new models due to the high performance demonstrated on the gold-standard data. The BioBERT and BioClinicalBERT were also used. The difference in performance was minimal compared to the base large uncased BERT model, thus, the metrics for these are not included. In addition, multiple combinations of dataset sampling were tried, primarily focusing on the number of samples for non-relations, as well as concept type filtering. It is necessary to mention that at the time of experimentation the number of relations annotated by the specialized team is not on a similar scale to the n2c2 dataset and that the datasets are specialty and task dependent, the Spatial dataset representing \textbf{0.84\%} and the Physiotherapy-Mobility representing \textbf{0.80\%} of n2c2's ~72000 relations.

Although the two datasets annotation tasks are different, `Spatial' and `Physiotherapy-Mobility', the results are comparable, as shown in Table \ref{tab:bert_spatial_base} , Table \ref{tab:bert_physio_base} respectively. We observe a performance increase when using the BERT based n2c2 models, an \textbf{16\%} overall increase for Spatial and a \textbf{15\%} increase for the Physiotherapy-Mobility model across the two classes when unfreezing layers. Similarly when unfreezing, the base BERT models have shown \textbf{11.5\%} increase in performance for Spatial and \textbf{10.8\%} for Physiotherapy-Mobility. Notably, Llama models maintain high performance even with all layers frozen. In this configuration, Spatial relations achieve the best scores among all its models.

\begin{enumerate}[label=\alph*.]
\item {Spatial dataset}
The four BERT model variants trained for this dataset show substantial differences, and although this is a binary classification problem, each relation task has many possible classes with sometimes as few as single digit samples for a given class. This is visible Appendix Table \ref{tab:spatialdatasettypes} where the relation class consists primarily of \textbf{body structures - disorders/abnormalities} and vice-versa while the `Other' negative class is comprised of mostly \textbf{qualifier values - body structures/qualifier values}.

The frozen BERT model showed imbalanced performance with a \textbf{20\%} precision-recall gap, suggesting poor identification of non-relations. The unfrozen model improved overall with a 0.90 F1-score and reduced precision-recall gap to \textbf{10\%}. While the frozen n2c2 BERT performed \textbf{3\%} worse than base BERT, the variant pre-trained with unfrozen layers achieved the best results, reaching 0.91 F1-score and 0.96 precision on Spatial class.

Llama with all layers frozen achieved performance of 0.93 across the two classes while still maintaining a similar gap of $\sim$10\% for precision and recall.

\vspace{0.5mm}
\item {Physiotherapy-Mobility dataset}
Several experiments were conducted with this dataset, the best results were achieved by increasing the number of non-relations. We observe that a considerable number of relations contain unknown types. In addition, the predominant non-relations are categorically different from the true class. Details available in Appendix Table \ref{tab:singleinstancedatasettypes}.

Increasing the proportion of non-relations improved results, likely due to the prevalence of unknown relation types. The fully frozen BERT model performed well, but had lower precision on non-relations. The unfrozen BERT model was more consistent, achieving a 0.90 F1-score. The standout was the fine-tuned n2c2 BERT with unfrozen layers, which reached a 0.93 F1-score - a \textbf{15\%} improvement over its frozen variant and \textbf{3\%} over the unfrozen base BERT. Llama also outperformed the frozen BERT at 0.835 F1-score, but remained behind the other BERT variants.

\end{enumerate}

\vspace{1mm}
\subsubsection{Performance of LLMs} The Mistral and Llama models show varying levels of classification performance in zero-shot and few-shot learning, as shown in Table \ref{tab:in_context_n2c2}. In zero-shot learning, the Mistral model shows particularly poor recall for the Strength-Drug, Dosage-Drug, and Form-Drug classes.  Overall, the Mistral model's performance is subpar, highlighting significant room for improvement. \newline
Similarly, the Llama model shows very low recall values for the Strength-Drug and Reason-Drug classes. For the minority classes, it achieves recall of 0.94 for ADE-Drug but only a moderate 0.56 for Duration-Drug. While the Llama model outperforms the Mistral model overall, its performance remains poor in comparison to our alternative approaches.

For few-shot learning, the Mistral model shows improvement, with the most significant gain observed for the Strength-Drug class, where although the recall improved by 30x, the resulting recall value of 0.21 remains poor. While half of the classes achieve good recall values, the other half continue to exhibit low performance. \newline
The Llama model also shows improved performance, with the minority classes achieving the highest recall values, with 0.95 for ADE-Drug and 0.82 for Duration-Drug. However, the model continues to struggle with three classes that exhibit poor performance, resulting in an overall sub-optimal outcome despite the improvement in some areas.

\begin{table} [h]
\caption{Performance summary for n2c2 dataset \\(\textit{For BERT and its variants, all layers are unfrozen; for Llama, the last layer is unfrozen)}}
\label{tab:perfsummaryn2c2}
\centering

 \begin{tabular}{ccccc}
%\begin{tabular}{p{2cm} p{1cm} p{1cm} p{1cm} p{1cm}}
\toprule
\multirow{2}{*}{Model} & \multirow{2}{*}{F1-score} & \multirow{2}{*}{Accuracy} & \multicolumn{2}{c}{Minority class F1-score} \\
\cline{4-5}
& & & ADE-D & Duration-D \\
\midrule

    BERT & 0.977 & 0.977 & 0.866 & 0.933 \\
    BioBERT & 0.955 & 0.955 & 0.742 & 0.894\\
    \makecell{BioClinical-\\BERT} & 0.942 & 0.942 & 0.693 & 0.857\\
    Llama & 0.949 & 0.96 & 0.626 & 0.9\\

\end{tabular}
\end{table}

%%%%%%%%%%%%%%%%%%%%%%%%%%%
\begin{table} [h]
\caption{Performance summary for Spatial dataset}
\label{tab:perfsummarynhs}
\centering

\begin{tabular}{ccc}
\toprule
    Model & F1-score & Accuracy\\
\midrule

    BERT (layers unfrozen) & 0.902 & 0.902 \\
    n2c2 trained BERT (layers unfrozen) & 0.918 & 0.918 \\
    Llama (layers frozen) & 0.933 & 0.933 \\

\end{tabular}
\end{table}

%%%%%%%%%%%%%%%%%%%%%%%%%%%
\begin{table} [h]
\caption{Performance summary for Physiotherapy-Mobility dataset}
\label{tab:perfsummarynhs2}
\centering

\begin{tabular}{ccc}
\toprule
    Model & F1-score & Accuracy\\
\midrule

    BERT (layers unfrozen) & 0.905 & 0.905 \\
    n2c2 trained BERT (layers unfrozen) & 0.938 & 0.938 \\
    Llama (layers frozen) & 0.835 & 0.835 \\

\end{tabular}
\end{table}

\section{Discussion}
\subsection{Performance on n2c2 dataset} 
BioBERT and BioClinicalBERT were outperformed by both BERT and Llama, which could be considered surprising given that these medical models are specifically trained on biomedical data. Also, these models perform much better with layers unfrozen, especially BioClinicalBERT. The cause for this can be explained below:
\begin{enumerate} [label=\alph*.]
    \item With the layers frozen, the models do not have the ability to learn as much and are forced to rely on their pre-training, restricting their capacity to adapt to new tasks. 
    \item After their pre-training, the models may have suffered from catastrophic forgetting \cite{DBLP:journals/corr/abs-2106-02902}, especially BioClinicalBERT as it had one additional training round; it was pre-trained on PubMed, PMC (same as BioBERT) and then additionally on MIMIC-III. This repeated exposure to specific biomedical texts, while potentially beneficial for certain tasks, might have caused the model to lose some of the generalization abilities it acquired during earlier stages of pre-training.
\end{enumerate}
When the layers are unfrozen, the models can learn and adapt more effectively. Their pre-training becomes an advantage rather than a limitation, as it provides a strong foundation that can be refined to adapt to the given task. 

Unfrozen BERT outperforms all other models. We attribute this to (base) BERT not having undergone additional pre-training beyond its initial phase allowing it to maintain a high degree of generalizability and a robust understanding of language.

With layers frozen, the Llama model achieves the highest overall performance among all models, demonstrating its effectiveness in capturing the nuances of the data with an F1-score of 0.929. 

With the last layer of the Llama model unfrozen, performance improves for the minority classes, similar to other models. However, the recall increase is subtle. We believe the model is overfitting, indicated by recall values nearing 1 during training. While unfreezing layers helps capture complex patterns for minority classes, it also increases overfitting, suggesting the need for regularization strategies. The model's substantial size further suggests that more data may be needed to prevent overfitting and enhance generalization.

\subsection{Performance on the NHS datasets}
 It is no surprise that the availability of samples plays an important role for each task and the number of true relations is low compared to the n2c2 dataset. One of the defining factors for performance improvements has been the adjustment of negative samples. Moreover, taking into account the generalization of the 'Spatial' and 'Physiotherapy-Mobility' class definition and concept sparsity we consider that on both datasets the models trained show acceptable performance, achieving over 0.9 scores across all metrics.
When using BERT fine-tuned on n2c2 as base we kept the new layers frozen with the hopes that whatever was learned in the previous models would be of use in the new task.
 
\begin{enumerate}[label=\alph*.]
\item {Spatial dataset}
For the Spatial classification task, given the base BERT model with frozen layers, the performance is reasonable, 0.79 F1. However, BERT fully unfrozen gives the model the necessary tunable parameters with the limited data, resulting in improvements in performance;although this could also signal overfitting.
The n2c2 fine-tuned BERT variants show marginally improved  performance in the case of the base unfrozen layer version, where the Spatial class registered an increase resulting in \textbf{0.91} F1-score, \textbf{0.90} accuracy and \textbf{0.87} recall, the previous knowledge from n2c2 being utilized. The frozen alternative proving to be of less use, likely due to the model weights not adapting to the variance of samples.

The fully frozen Llama model has the best performance overall, evidencing its adaptability to the task compared to the base frozen BERT variants while maintaining consistency for both classes, ~0.93 F1-Score.

\vspace{0.5mm}
\item {Physiotherapy-Mobility dataset}

The 'Single Instance' class has initially proven to be hard to classify due to the identity of the relation, it resembles a context wide concept which can be hard to identify even by the annotators. 

BERT with frozen layers has difficulty in identifying the true class with ~0.66 precision. The unfrozen alternative brings a substantial improvement in accuracy and F1-core and recall, it being of \textbf{0.90} and \textbf{0.89} respectively, signaling that the task given is hard. The n2c2 fine-tuned BERT models have performed slightly better than the aforementioned base models.
 
Llama with frozen layers has shown marginally better performance than both its base BERT \& BERT n2c2 frozen counterparts, an overall improvement of \textbf{3-5\%} across all metrics.

\end{enumerate}

\section{Limitations \& Future work}
While we have shown that both BERT and Llama are performing well in our use cases for inter-entity relation extraction or classification, further developments can be explored. In our work, the datasets and relations considered have been labeled and all the training supervised. However, by leveraging an ontology dictionary such as SNOMED CT and the NER capabilities of MedCAT, we can attempt to create labels based only valid entity types, such examples could be "clinical drug - dose form" or "clinical drug - event". Building on the notion of relation discovery. 

We have integrated the model training pipeline, annotation tool and and inference pipeline openly available integrating them within the MedCAT library, allowing the community to replicate our results and expand models to new relation types.

\section{Conclusion}

In conclusion, RelCAT represents a significant advancement in clinical natural language processing by providing a comprehensive toolkit for relation concept annotation. By leveraging the CogStack MedCAT framework and integrating advanced machine learning models like BERT and Llama, RelCAT effectively addresses the complex challenge of extracting and classifying relations between entities in clinical narratives. The tool-kit's innovative approach was validated through rigorous experimentation, demonstrating remarkable performance with a macro F1-score and recall of 0.977 on the gold-standard n2c2 dataset, which notably surpasses previous state-of-the-art benchmarks. Moreover, RelCAT's consistent performance across both gold-standard and real-world NHS clinical datasets, BERT obtaining an F1 of \textbf{~0.94} and recall \textbf{~0.94} for the Physiotherapy dataset, and for the Spatial dataset Llama3 achieving an F1 and recall \textbf{0.93} and \textbf{0.93} respectively, which underscores its robustness and potential to transform clinical text analysis. This work not only introduces an update to an already powerful annotation tool but also provides a versatile framework for researchers and clinicians seeking to unlock meaningful insights from complex medical text data. The complete source code for RelCAT is freely available through a public GitHub repository at https://github.com/CogStack/MedCAT. The repository contains all necessary components to test and deploy the described methods. The code repository is actively maintained and includes all implementations necessary to reproduce the results presented in this paper as of January 2025.

\section*{Acknowledgments}
We appreciate the help and support from GSTT, KCH, the GSTT-Cogstack team, the KCH-Cogstack team, James Teo, Dan Stein, Christian Godfrey and Nathan Beniach for helping with data collection in this work.
This work was supported by Health Data Research UK, an initiative funded by UK Research and Innovation, Department of Health and Social Care
(England) and the devolved administrations, and
leading medical research charities. SA, VD, TS, MR, RD are part-funded by the National Institute for Health Research (NIHR) Biomedical
Research Centre at South London and Maudsley
NHS Foundation Trust and King’s College
London. RD is also supported by The National Institute
for Health Research University College London
Hospitals Biomedical Research Centre.
This paper represents independent research
part funded by the National Institute for Health
Research (NIHR) Biomedical Research Centre at
South London and Maudsley NHS Foundation
Trust and King’s College London. The views expressed are those of the authors and not necessarily
those of the NHS, the NIHR or the Department of
Health and Social Care. The funders had no role in
study design, data collection and analysis, decision
to publish, or preparation of the manuscript.
\onecolumn

\section{Appendix}

\noindent \textit{Example of the LLM prompt used for zero and few shot}
\newline \newline
Prompt for Llama-3.1-8B-Instruct model
\begin{lstlisting}
    """
    <|begin_of_text|><|start_head_id|>system<|end_header_id|>
    You are a relationship classification bot. 
    Your task is to assess intent and categorize the relationship between the entities in the input text into one of the predefined categories.
    You will only respond with the predefined category. 
    Do not provide explanations or notes. 
    <|eot_id|> <|start_head_id|>user<|end_header_id|>
    CLassify the relationships into the following categories:
    0: 'Reason-Drug', 1: 'Duration-Drug', 2: 'ADE-Drug', 3: 'Dosage-Drug', 4: 'Strength-Drug', 5: 'Route-Drug', 6: 'Frequency-Drug', 7: 'Form-Drug'

    Explanation of categories:
    0: relationship explains why the drug was prescribed
    1: relationship explains how long the drug was prescribed for
    2: relationship explains the adverse effect the drug had
    3: relationship explains the amount (dosage) of the prescribed drug
    4: relationship explains the strength of the prescribed drug
    5: relationship explains the route through which the drug is to be adminstered
    6: relationship explains the frequency of the prescribed drug
    7: relationship explains the given form of the prescribed drug

    Format of input: 'tokens', 'entity 1', 'entity 2'. Classify the relationship between the entities present in the tokens.
    Input: {input} <|eot_id|>
    <|start_header_id|>assistant<|end_header_id|>
    """


\end{lstlisting}

Prompt for Mistral-7B-Instruct-v0.2 model
\begin{lstlisting}
    """
    <s>[INST]You are a relationship classification bot. 
    Your task is to assess intent and categorize the relationship between the entities in the input text into one of the predefined categories.
    You will only respond with the predefined category. 
    Do not provide explanations or notes. 
    
    ###
    Classify the relationships into the following categories:
    0: 'Reason-Drug', 1: 'Duration-Drug', 2: 'ADE-Drug', 3: 'Dosage-Drug', 4: 'Strength-Drug', 5: 'Route-Drug', 6: 'Frequency-Drug', 7: 'Form-Drug'

    Explanation of categories:
    0: relationship explains why the drug was prescribed
    1: relationship explains how long the drug was prescribed for
    2: relationship explains the adverse effect the drug had
    3: relationship explains the amount (dosage) of the prescribed drug
    4: relationship explains the strength of the prescribed drug
    5: relationship explains the route through which the drug is to be adminstered
    6: relationship explains the frequency of the prescribed drug
    7: relationship explains the given form of the prescribed drug

    Format of input: 'tokens', 'entity 1', 'entity 2'. Classify the relationship between the entities present in the tokens.
    
    ###
    Input: {input}[/INST]
    """


\end{lstlisting}

\begin{table*} [h!]
\caption{Relation type distribution (top 10) NHS Spatial dataset by class}
\label{tab:spatialdatasettypes}
\centering

\begin{tabular}{m{3cm} m{5cm} m{4cm}} 
\toprule
    Class / type & Number of relations by concept type pair (t1-t2)\\
\midrule

 Spatial & body structure-morphologic abnormality & 37\\
 & body structure-disorder & 35\\
 & morphologic abnormality-body structure & 27\\
 & body structure-finding & 21\\
 & disorder-body structure & 21\\
 & finding-body structure & 16\\
 & body structure-body structure & 16\\
 & qualifier value-body structure & 11\\
 & body structure-procedure & 7\\
\midrule
 Other & qualifier value-qualifier value & 76\\
 & qualifier value-body structure & 21\\
 & body structure-qualifier value & 13\\
 & procedure-qualifier value & 12\\
 & qualifier value-disorder & 12\\
 & qualifier value-morphologic abnormality & 10\\
 & qualifier value-finding & 10\\
 & disorder-qualifier value & 9\\
 & finding-qualifier value & 8\\
 & qualifier value-procedure & 6\\

\end{tabular}
\end{table*}

\vspace{0.5mm}

\begin{table*} [h]
\caption{Relation type distribution (top 10) NHS Physiotherapy-Mobility dataset by class}
\label{tab:singleinstancedatasettypes}
\centering

%\begin{tabular}{m{3cm} m{5cm} m{4cm}}
%\toprule
%    Class / type & Number of relations by concept type pair (t1-t2)\\
%\midrule
%
%    Single Instance & finding-undefined & 56 \\
%        & qualifier value-procedure &46 \\
%        & undefined-procedure       &34 \\
%        & undefined-finding         &19 \\
%        & finding-finding & 9  \\
%        & procedure-qualifier value &6 \\
%        & procedure-undefined       &6 \\
%        & record artifact-procedure &5 \\
%        & physical object-undefined &1 \\
%        & assessment scale-undefined &1 \\
%    \midrule
%    Other  & finding-finding & 129 \\
%        & finding-undefined & 38 \\
%        & undefined-finding & 27 \\
%        & undefined-undefined & 15 \\
%        & qualifier value-finding & 16 \\
%        & finding-event & 10 \\
%        & event-finding & 9 \\
%        & finding-qualifier value & 8 \\
%        & procedure-finding & 7 \\
%        & qualifier value-qualifier value & 7 \\
%
%
%\end{tabular}
%\end{table*}

\begin{tabular}{m{3cm} m{5cm} m{4cm}}
\toprule
    Class / type & Number of relations by concept type pair (t1-t2)\\
\midrule

    Single Instance & finding-procedure & 139 \\
        & qualifier value-procedure & 71 \\
        & procedure-finding & 39 \\
        & finding-finding & 13 \\
        & procedure-qualifier value & 9 \\
        & record artifact-procedure & 5 \\
        & physical object-procedure & 1 \\
        & assessment scale-procedure & 1 \\

    \midrule
    Other  & finding-finding & 279 \\
        & finding-procedure & 67 \\
        & procedure-finding & 58 \\
        & qualifier value-finding & 31 \\
        & finding-qualifier value & 25 \\
        & finding-event & 23 \\
        & qualifier value-qualifier value & 21\\
        & qualifier value-procedure & 17 \\
        & procedure-procedure & 13\\
        & event-finding & 10 \\

\end{tabular}
\end{table*}

%%%%%%%%%%%%%%%%%%%%%

\begin{table*}[h]
\centering
\caption{BERT model performance for n2c2 dataset}
\label{tab:bert_n2c2}
\begin{tabular}{m{2cm} | m{1cm} m{1cm} m{0.8cm} m{1cm} | m{1cm} m{1cm} m{0.8cm} m{1cm} } 

\toprule
\multirow{2}{*}{Class} & \multicolumn{4} {c|} {\textbf{Layers frozen}} & \multicolumn{4} {c} {\textbf{Layers unfrozen}} \\
 & Accuracy & F1-score & Recall & Precision & Accuracy & F1-score & Recall & Precision \\

\midrule

Frequency-Drug & 0.981 & 0.942 & 0.955 & 0.932  & 0.998 & 0.993 & 0.995 & 0.991 \\ 
Strength-Drug  & 0.982 & 0.950 & 0.967 & 0.935 & 0.996 & 0.990 & 0.992 & 0.987 \\ 
Reason-Drug    & 0.932 & 0.721 & 0.841 & 0.636 & 0.990 & 0.964 & 0.954 & 0.974 \\ 
Route-Drug     & 0.960 & 0.869 & 0.845 & 0.897 & 0.992 & 0.975 & 0.974 & 0.976 \\ 
Duration-Drug  & 0.974 & 0.760 & 0.666 & 0.931 & 0.982 & 0.933 & 0.924 & 0.954 \\ 
Dosage-Drug    & 0.975 & 0.891 & 0.896 & 0.891 & 0.995 & 0.977 & 0.974 & 0.982\\ 
Form-Drug      & 0.975 & 0.930 & 0.971 & 0.893 & 0.992 & 0.979 & 0.986 & 0.974\\ 
ADE-Drug       & 0.946 & 0.479 & 0.348 & 0.832 & 0.993 & 0.866 & 0.890 & 0.861 \\ 

\midrule
 \multicolumn{1}{c}{\textbf{\makecell{}}} & \multicolumn{1} {c}{\textbf{\makecell{Accuracy: \\ 0.871}}} & \textbf{\makecell{F1-score:\\ 0.871}} & \textbf{\makecell{Recall:\\ 0.871}} & \multicolumn{1} {c|}{\textbf{\makecell{Precision:\\ 0.871}}} & \multicolumn{1} {c}{\textbf{\makecell{Accuracy: \\ 0.977}}} & \textbf{\makecell{F1-score:\\ 0.977}} & \textbf{\makecell{Recall:\\ 0.977}} & \multicolumn{1} {c}{\textbf{\makecell{Precision:\\ 0.977}}}\\
 
\end{tabular}
\end{table*}

\begin{table*}[h]
\centering
\caption{BioBERT model performance for n2c2 dataset}
\label{tab:biobert_n2c2}
\begin{tabular}{m{2cm} | m{1cm} m{1cm} m{0.8cm} m{1cm} | m{1cm} m{1cm} m{0.8cm} m{1cm} } 

\toprule
\multirow{2}{*}{Class} & \multicolumn{4} {c|} {\textbf{Layers frozen}} & \multicolumn{4} {c} {\textbf{Layers unfrozen}} \\
 & Accuracy & F1-score & Recall & Precision & Accuracy & F1-score & Recall & Precision \\

\midrule

Frequency-Drug & 0.939 & 0.820 & 0.818 & 0.829  & 0.993 & 0.980 & 0.986 & 0.975 \\ 
Strength-Drug  & 0.956 & 0.870 & 0.918 & 0.832 & 0.994 & 0.982 & 0.980 & 0.985 \\ 
Reason-Drug    & 0.899 & 0.602 & 0.672 & 0.558 & 0.977 & 0.914 & 0.932 & 0.900 \\ 
Route-Drug     & 0.950 & 0.822 & 0.873 & 0.783 & 0.986 & 0.955 & 0.940 & 0.972 \\ 
Duration-Drug  & 0.874 & 0.410 & 0.334 & 0.629 & 0.996 & 0.894 & 0.877 & 0.937 \\ 
Dosage-Drug    & 0.952 & 0.793 & 0.771 & 0.827 & 0.992 & 0.966 & 0.956 & 0.978\\ 
Form-Drug      & 0.958 & 0.878 & 0.938 & 0.829 & 0.988 & 0.967 & 0.977 & 0.958 \\ 
ADE-Drug       & 0.910 & 0.332 & 0.241 & 0.613 & 0.984 & 0.724 & 0.742 & 0.735 \\ 

\midrule
 \multicolumn{1}{c}{\textbf{\makecell{}}} & \multicolumn{1} {c}{\textbf{\makecell{Accuracy: \\ 0.776}}} & \textbf{\makecell{F1-score:\\ 0.777}} & \textbf{\makecell{Recall:\\ 0.779}} & \multicolumn{1} {c|}{\textbf{\makecell{Precision:\\ 0.776}}} & \multicolumn{1} {c}{\textbf{\makecell{Accuracy: \\ 0.955}}} & \textbf{\makecell{F1-score:\\ 0.955}} & \textbf{\makecell{Recall:\\ 0.955}} & \multicolumn{1} {c}{\textbf{\makecell{Precision:\\ 0.955}}}\\
 
\end{tabular}
\end{table*}

\begin{table*}[h]
\centering
\caption{BioClinicalBERT model performance for n2c2 dataset with all layers frozen}
\label{tab:bioclinicalbert_n2c2}
\begin{tabular}{m{2cm} | m{1cm} m{1cm} m{0.8cm} m{1cm} | m{1cm} m{1cm} m{0.8cm} m{1cm} } 

\toprule
\multirow{2}{*}{Class} & \multicolumn{4} {c|} {\textbf{Layers frozen}} & \multicolumn{4} {c} {\textbf{Layers unfrozen}} \\
 & Accuracy & F1-score & Recall & Precision & Accuracy & F1-score & Recall & Precision \\

\midrule

Frequency-Drug & 0.926 & 0.770 & 0.832 & 0.719 & 0.990 & 0.970 & 0.973 & 0.967 \\ 
Strength-Drug  & 0.948 & 0.848 & 0.904 & 0.802 & 0.991 & 0.975 & 0.974 & 0.976 \\ 
Reason-Drug    & 0.875 & 0.491 & 0.577 & 0.433 & 0.972 & 0.898 & 0.898 & 0.900 \\ 
Route-Drug     & 0.925 & 0.738 & 0.785 & 0.700 & 0.984 & 0.948 & 0.946 & 0.951 \\ 
Duration-Drug  & 0.949 & 0.376 & 0.271 & 0.684 & 0.996 & 0.857 & 0.892 & 0.850 \\ 
Dosage-Drug    & 0.942 & 0.754 & 0.734 & 0.785 & 0.986 & 0.941 & 0.927 & 0.957\\ 
Form-Drug      & 0.952 & 0.863 & 0.911 & 0.822 & 0.984 & 0.955 & 0.967 & 0.945 \\ 
ADE-Drug       & 0.903 & 0.297 & 0.195 & 0.685 & 0.982 & 0.693 & 0.702 & 0.713 \\ 

\midrule
 \multicolumn{1}{c}{\textbf{\makecell{}}} & \multicolumn{1} {c}{\textbf{\makecell{Accuracy: \\ 0.716}}} & \textbf{\makecell{F1-score:\\ 0.716}} & \textbf{\makecell{Recall:\\ 0.716}} & \multicolumn{1} {c|}{\textbf{\makecell{Precision:\\ 0.716}}} & \multicolumn{1} {c}{\textbf{\makecell{Accuracy: \\ 0.942}}} & \textbf{\makecell{F1-score:\\ 0.942}} & \textbf{\makecell{Recall:\\ 0.942}} & \multicolumn{1} {c}{\textbf{\makecell{Precision:\\ 0.942}}}\\
 
\end{tabular}
\end{table*}

%%%%%%%%%%%%%%%%%%%%%%%%%%%%%%%%%%%%%%%

%%%%%%%%%%%%%%%%%%%%%%%%%%%%%
\begin{table*}[h]
\centering
\caption{Llama Model performance for n2c2 dataset}
\label{tab:llama_n2c2}
\begin{tabular}{m{2cm} | m{1cm} m{1cm} m{0.8cm} m{1cm} | m{1cm} m{1cm} m{0.8cm} m{1cm} } 

\toprule
\multirow{2}{*}{Class} & \multicolumn{4} {c|} {\textbf{Layers frozen}} & \multicolumn{4} {c} {\textbf{Layer unfrozen}} \\
 & Accuracy & F1-score & Recall & Precision & Accuracy & F1-score & Recall & Precision \\

\midrule
Frequency-Drug &  0.99  & 0.97 & 0.97 & 0.972  &  0.993  & 0.979 & 0.982 & 0.977 \\ 
Strength-Drug  & 0.99 & 0.971 & 0.976 & 0.967 & 0.994 &  0.982 & 0.987 & 0.978 \\ 
Reason-Drug    & 0.958 & 0.857 & 0.806 & 0.923 & 0.971 &  0.9 & 0.857 & 0.952 \\ 
Route-Drug     & 0.983 & 0.943 & 0.947 & 0.942 & 0.987 & 0.956 & 0.985 & 0.93 \\ 
Duration-Drug  &  0.887 & 0.664 & 0.758 & 0.62 & 0.985 & 0.9 & 0.924 & 0.792 \\ 
Dosage-Drug    & 0.987 & 0.944 & 0.944 & 0.949 & 0.991 & 0.961 & 0.963 & 0.960\\ 
Form-Drug      & 0.984 & 0.957 & 0.961 & 0.955 & 0.986 & 0.962 & 0.95 & 0.974 \\ 
ADE-Drug       & 0.956 & 0.386  & 0.539 & 0.337 & 0.991 & 0.626  & 0.713 & 0.59 \\ 

\midrule
\multicolumn{1}{c}{\textbf{\makecell{}}} & \multicolumn{1} {c}{\textbf{\makecell{Accuracy: \\ 0.95}}} & \textbf{\makecell{F1-score:\\ 0.929}} & \textbf{\makecell{Recall:\\ 0.929}} & \multicolumn{1} {c|}{\textbf{\makecell{Precision:\\ 0.929}}} & \multicolumn{1} {c}{\textbf{\makecell{Accuracy: \\ 0.96}}} & \textbf{\makecell{F1-score:\\ 0.949}} & \textbf{\makecell{Recall:\\ 0.949}} & \multicolumn{1} {c}{\textbf{\makecell{Precision:\\ 0.949}}}\\ 
\end{tabular}
\end{table*}

%%%%%%%%%%%%%%%%%%%%%%%%%%%%%%%%%%%%%%%
\begin{table*}[h]
\centering
\caption{BERT model performance for n2c2 dataset with last layer unfrozen}
\label{tab:bert_lastu_n2c2}
\begin{tabular}{m{2cm} m{1cm} m{0.8cm} m{0.8cm} m{0.8cm} }

\toprule
Class & Accuracy & F1-score & Recall & Precision \\

\midrule

Frequency-Drug & 0.998 & 0.991  & 0.998 & 0.991\\
Strength-Drug & 0.996  & 0.988 & 0.992 & 0.986\\
Reason-Drug & 0.986  & 0.94 & 0.932 & 0.97\\
Route-Drug & 0.992  & 0.973 & 0.979 & 0.969\\
Duration-Drug & 0.987 & 0.922 & 0.937 & 0.946\\
Dosage-Drug & 0.994 & 0.972 & 0.97 & 0.979\\
ADE-Drug & 0.984 & 0.77 & 0.866 & 0.792\\
Form-Drug & 0.992 & 0.977 & 0.98 & 0.978\\

\bottomrule
 
\end{tabular}
\end{table*}

%%%%%%%%%%%%%%%%%%%%%%%%%%%%%%%%%%%%%%%
\begin{table*}[h]
\centering
\caption{BERT model performance for n2c2 dataset with last layer frozen}
\label{sub-type}
\begin{tabular}{m{2cm} m{1cm} m{0.8cm} m{0.8cm} m{0.8cm} }

\toprule
Class & Accuracy & F1-score & Recall & Precision \\

\midrule

Frequency-Drug  & 0.998 & 0.994 & 0.996 & 0.992 \\
Strength-Drug   & 0.997 & 0.991 & 0.997 & 0.985 \\
Reason-Drug     & 0.990 & 0.963 & 0.965 & 0.962 \\
Route-Drug      & 0.992 & 0.973 & 0.967 & 0.979 \\
Duration-Drug   & 0.982 & 0.935 & 0.930 & 0.951 \\
Dosage-Drug     & 0.995 & 0.978 & 0.969 & 0.988 \\
ADE-Drug        & 0.992 & 0.873 & 0.857 & 0.911 \\
Form-Drug       & 0.992 & 0.979 & 0.987 & 0.971 \\

\midrule
\textbf{\makecell{}} & \textbf{\makecell{Accuracy: \\ 0.977}} & \textbf{\makecell{F1-score:\\ 0.977}} & \textbf{\makecell{Recall:\\ 0.977}} & \textbf{\makecell{Precision:\\ 0.977}}\\
 
\end{tabular}
\end{table*}

%%%%%%%%%%%%%%%%%%%%%%%

\begin{table*}
\centering
\caption{LLM (in-context learning) performance for n2c2 dataset}
\label{tab:in_context_n2c2}
%\begin{tabular}{ |m{2cm}|m{2.5cm}|m{2cm}|m{3cm}|m{2cm}|m{3cm}|} 
\begin{tabular}{ c c c c } 

\toprule
\multirow{2}{*}{Class} & \multirow{2}{*}{Model} & Zero-shot & Few-shot\\
 & & Recall & Recall \\

\midrule
\multirow{2}{*}{Frequency-Drug}  & Llama & 0.83 & 0.81 \\
& Mistral & 0.83 & 0.89\\

\midrule

\multirow{2}{*}{Strength-Drug}  & Llama  & 0.001  & 0.05 \\
& Mistral  & 0.007  & 0.21 \\

\midrule

\multirow{2}{*}{Reason-Drug}  & Llama  & 0.02  & 0.05 \\
& Mistral  & 0.286  & 0.72 \\

\midrule

\multirow{2}{*}{Route-Drug}  & Llama  & 0.51  & 0.3 \\
& Mistral  & 0.5  & 0.62 \\

\midrule

\multirow{2}{*}{Duration-Drug}  & Llama  & 0.56  & 0.82 \\
& Mistral  & 0.61  & 0.62 \\

\midrule

\multirow{2}{*}{Dosage-Drug}  & Llama  & 0.51  & 0.64 \\
& Mistral  & 0.01  & 0.01 \\

\midrule

\multirow{2}{*}{ADE-Drug}  & Llama  & 0.94  & 0.95 \\
& Mistral  & 0.44  & 0.36 \\

\midrule

\multirow{2}{*}{Form-Drug}  & Llama  & 0.53  & 0.55 \\
& Mistral  & 0.01  & 0.13 \\

\midrule

\textbf{Mistral} & \textbf{Zero shot} & \textbf{\makecell{Accuracy: \\ 0.29}} & \textbf{\makecell{F1-score: \\ 0.32}} \\
\textbf{Mistral} & \textbf{Few shot} & \textbf{\makecell{Accuracy: \\ 0.35}} & \textbf{\makecell{F1-score: \\ 0.4}} \\
\textbf{Llama} & \textbf{Zero shot} & \textbf{\makecell{Accuracy: \\ 0.47}} & \textbf{\makecell{F1-score: \\ 0.49}} \\
\textbf{Llama} & \textbf{Few shot} & \textbf{\makecell{Accuracy: \\ 0.41}} & \textbf{\makecell{F1-score: \\ 0.46}} \\ \\
 
\end{tabular}
\end{table*}

%%%%%%%%%%%%%%%%%%

\begin{table*}[h]
\centering
\caption{BERT performance for Spatial dataset}
\label{tab:bert_spatial_base}
\begin{tabular}{m{2cm} | m{1cm} m{1cm} m{0.8cm} m{1cm} | m{1cm} m{1cm} m{0.8cm} m{1cm} } 

\toprule
\multirow{2}{*}{Class} & \multicolumn{4} {c|} {\textbf{Layers frozen}} & \multicolumn{4} {c} {\textbf{Layers unfrozen}} \\
 & Accuracy & F1-score & Recall & Precision & Accuracy & F1-score & Recall & Precision \\

\midrule
Spatial         & 0.787 & 0.794 & 0.714 & 0.893 & 0.902 & 0.900 & 0.844 & 0.964\\
Other           & 0.787 & 0.780 & 0.885 & 0.697 & 0.902 & 0.903 & 0.966 & 0.848\\
\midrule
\multicolumn{1}{c}{\textbf{\makecell{}}} & \multicolumn{1} {c}{\textbf{\makecell{Accuracy: \\ 0.787}}} & \textbf{\makecell{F1-score:\\ 0.787}} & \textbf{\makecell{Recall:\\ 0.787}} & \multicolumn{1} {c|}{\textbf{\makecell{Precision:\\ 0.787}}} & \multicolumn{1} {c}{\textbf{\makecell{Accuracy: \\ 0.902}}} & \textbf{\makecell{F1-score:\\ 0.902}} & \textbf{\makecell{Recall:\\ 0.902}} & \multicolumn{1} {c}{\textbf{\makecell{Precision:\\ 0.902}}}\\
 
\end{tabular}
\end{table*}

%%%%%%%%%%%%%%%%%%%%%%%%%%%%%%%%%%%%%%%

% \begin{table}[h]
% \centering
% \caption{BERT performance for Spatial dataset with all layers unfrozen}
% \label{sub-type}
% \begin{tabular}{m{2cm} m{1cm} m{0.8cm} m{0.8cm} m{0.8cm} } 

% \toprule
% Class & Accuracy & F1-score & Recall & Precision \\

% \midrule
% Spatial         & 0.902 & 0.900 & 0.844 & 0.964\\
% Other           & 0.902 & 0.903 & 0.966 & 0.848\\
% \midrule
%  \textbf{\makecell{}} & \textbf{\makecell{Accuracy: \\ 0.902}} & \textbf{\makecell{F1-score:\\ 0.902}} & \textbf{\makecell{Recall:\\ 0.902}} & \textbf{\makecell{Precision:\\ 0.902}}\\
 
% \end{tabular}
% \end{table}
%%%%%%%%%%%%%%%%%%%%%%%%%%%%%%%%%%%%%%%

\begin{table*}[h]
\centering
\caption{n2c2 trained BERT performance for Spatial dataset}
\label{tab:bert_spatial_n2c2}
\begin{tabular}{m{2cm} | m{1cm} m{1cm} m{0.8cm} m{1cm} | m{1cm} m{1cm} m{0.8cm} m{1cm} } 

\toprule
\multirow{2}{*}{Class} & \multicolumn{4} {c|} {\textbf{Layers frozen}} & \multicolumn{4} {c} {\textbf{Layers unfrozen}} \\
 & Accuracy & F1-score & Recall & Precision & Accuracy & F1-score & Recall & Precision \\

\midrule
Spatial         & 0.750 & 0.787 & 0.704 & 0.893  & 0.918 & 0.915 & 0.871 & 0.964\\
Other           &0.750 & 0.697 & 0.838 & 0.596 & 0.918 & 0.921 & 0.967 & 0.879\\
\midrule
\multicolumn{1}{c}{\textbf{\makecell{}}} & \multicolumn{1} {c}{\textbf{\makecell{Accuracy: \\ 0.750}}} & \textbf{\makecell{F1-score:\\ 0.750}} & \textbf{\makecell{Recall:\\ 0.750}} & \multicolumn{1} {c|}{\textbf{\makecell{Precision:\\ 0.750}}} & \multicolumn{1} {c}{\textbf{\makecell{Accuracy: \\ 0.918}}} & \textbf{\makecell{F1-score:\\ 0.918}} & \textbf{\makecell{Recall:\\ 0.918}} & \multicolumn{1} {c}{\textbf{\makecell{Precision:\\ 0.918}}}\\
 
\end{tabular}
\end{table*}
%%%%%%%%%%%%%%%%

\begin{table*}[h]
\centering
\caption{Llama performance for Spatial dataset with all layers frozen}
\label{tab:llama_spatial_base}
\begin{tabular}{m{2cm} m{1cm} m{0.8cm} m{0.8cm} m{0.8cm} } 

\toprule
Class & Accuracy & F1-score & Recall & Precision \\

\midrule
Spatial         & 0.933 & 0.931 & 0.886 & 0.981 \\
Other           & 0.933 & 0.935 & 0.982 & 0.891 \\
\midrule
 \textbf{\makecell{}} & \textbf{\makecell{Accuracy: \\ 0.933}} & \textbf{\makecell{F1-score:\\ 0.933}} & \textbf{\makecell{Recall:\\ 0.933}} & \textbf{\makecell{Precision:\\ 0.933}}\\
 
\end{tabular}
\end{table*}

%%%%%%%%%%%%%%%%%%%%%%%%%%%%%%%%%%%%%%%

\begin{table*}[h]
\centering
\caption{BERT performance for Physiotherapy-Mobility dataset}
\label{tab:bert_physio_base}
\begin{tabular}{m{2cm} | m{1cm} m{1cm} m{0.8cm} m{1cm} | m{1cm} m{1cm} m{0.8cm} m{1cm} } 

\toprule
\multirow{2}{*}{Class} & \multicolumn{4} {c|} {\textbf{Layers frozen}} & \multicolumn{4} {c} {\textbf{Layers unfrozen}} \\
 & Accuracy & F1-score & Recall & Precision & Accuracy & F1-score & Recall & Precision \\

\midrule
Other           & 0.797 & 0.824 & 0.745 & 0.921 & 0.905 & 0.907 & 0.919 & 0.895\\
Single instance & 0.797 & 0.762 & 0.889 & 0.667 & 0.905 & 0.904 & 0.892 & 0.917\\
\midrule
\multicolumn{1}{c}{\textbf{\makecell{}}} & \multicolumn{1} {c}{\textbf{\makecell{Accuracy: \\ 0.797}}} & \textbf{\makecell{F1-score:\\ 0.797}} & \textbf{\makecell{Recall:\\ 0.797}} & \multicolumn{1} {c|}{\textbf{\makecell{Precision:\\ 0.797}}} & \multicolumn{1} {c}{\textbf{\makecell{Accuracy: \\ 0.905}}} & \textbf{\makecell{F1-score:\\ 0.905}} & \textbf{\makecell{Recall:\\ 0.905}} & \multicolumn{1} {c}{\textbf{\makecell{Precision:\\ 0.905}}}\\
 
\end{tabular}
\end{table*}
%%%%%%%%%%%%%%%%%%%%%%%%%%%%%%%%%
\begin{table*}[h]
\centering
\caption{n2c2 trained BERT performance for Physiotherapy-Mobility dataset}
\label{tab:bert_physio_n2c2}
\begin{tabular}{m{2cm} | m{1cm} m{1cm} m{0.8cm} m{1cm} | m{1cm} m{1cm} m{0.8cm} m{1cm} } 

\toprule
\multirow{2}{*}{Class} & \multicolumn{4} {c|} {\textbf{Layers frozen}} & \multicolumn{4} {c} {\textbf{Layers unfrozen}} \\
 & Accuracy & F1-score & Recall & Precision & Accuracy & F1-score & Recall & Precision \\

\midrule
Single instance & 0.784 & 0.712 & 0.722 & 0.703 & 0.938 & 0.923 & 0.870 & 0.973\\
Other           & 0.784 & 0.826 & 0.820 & 0.833 & 0.938 & 0.948 & 0.982 & 0.917\\
\midrule
\multicolumn{1}{c}{\textbf{\makecell{}}} & \multicolumn{1} {c}{\textbf{\makecell{Accuracy: \\ 0.784}}} & \textbf{\makecell{F1-score:\\ 0.784}} & \textbf{\makecell{Recall:\\ 0.784}} & \multicolumn{1} {c|}{\textbf{\makecell{Precision:\\ 0.784}}} & \multicolumn{1} {c}{\textbf{\makecell{Accuracy: \\ 0.938}}} & \textbf{\makecell{F1-score:\\ 0.938}} & \textbf{\makecell{Recall:\\ 0.938}} & \multicolumn{1} {c}{\textbf{\makecell{Precision:\\ 0.938}}}\\
 
\end{tabular}
\end{table*}
%%%%%%%%%%%%%%%%%

\begin{table*}
\centering
\caption{Llama performance for Physiotherapy-Mobility dataset with all layers frozen}
\label{tab:bert_physio_llama}
\begin{tabular}{m{2cm} m{1cm} m{0.8cm} m{0.8cm} m{0.8cm} } 

\toprule
Class & Accuracy & F1-score & Recall & Precision \\

\midrule
Single instance & 0.835 & 0.788 & 0.802 & 0.776 \\
Other           & 0.835 & 0.863 & 0.852 & 0.873 \\
\midrule
 \textbf{\makecell{}} & \textbf{\makecell{Accuracy: \\ 0.835}} & \textbf{\makecell{F1-score:\\ 0.835}} & \textbf{\makecell{Recall:\\ 0.835}} & \textbf{\makecell{Precision:\\ 0.835}}\\

\end{tabular}
\end{table*}

%%%%%%%%%%%%%%%%%%%%

\clearpage
\bibliographystyle{ieeetr}
\bibliography{dis}

\end{document}